\documentclass[]{spie}  

\usepackage{amsmath,amsfonts,amssymb}
\usepackage{graphicx}
\usepackage{subfigure}
\usepackage{multirow}
\usepackage{wrapfig}
\usepackage{booktabs}
\usepackage[colorlinks=true, allcolors=blue]{hyperref}

\usepackage{setspace}

\title{Unsupervised Generation of Pseudo Normal PET from MRI with Diffusion Model for Epileptic Focus Localization}

\author[a]{Wentao Chen}
\author[b]{Jiwei Li}
\author[c]{Xichen Xu}
\author[b]{Hui Huang}
\author[b]{Siyu Yuan}
\author[d]{Miao Zhang}
\author[c]{Tianming Xu}
\author[b]{Jie Luo}
\author[a,c]{Weimin Zhou}
\affil[a]{University of Michigan-Shanghai Jiao Tong University Joint Institute, Shanghai Jiao Tong University, Shanghai, 200240, China}
\affil[b]{School of Biomedical Engineering, Shanghai Jiao Tong University, Shanghai, 200240, China}
\affil[c]{Global Institute of Future Technology, Shanghai Jiao Tong University, Shanghai, 200240, China}
\affil[d]{Department of Nuclear Medicine, Ruijin Hospital, Shanghai Jiao Tong University School of Medicine, Shanghai, 200025, China}

\authorinfo{Further author information: (Send correspondence to Jie Luo and Weimin Zhou.)\\Jie Luo: E-mail: jieluo@sjtu.edu.cn\\
Weimin Zhou: E-mail: weimin.zhou@sjtu.edu.cn}

\begin{document} 
\maketitle

\begin{abstract}
[$^{18}$F]fluorodeoxyglucose (FDG) positron emission tomography (PET) has emerged as a crucial tool in identifying the epileptic focus, especially in cases where magnetic resonance imaging (MRI) diagnosis yields indeterminate results. FDG PET can provide the metabolic information of glucose and help identify abnormal areas that are not easily found through MRI. However, the effectiveness of FDG PET-based assessment and diagnosis depends on the selection of a healthy control group. The healthy control group typically consists of healthy individuals similar to epilepsy patients in terms of age, gender, and other aspects for providing normal FDG PET data, which will be used as a reference for enhancing the accuracy and reliability of the epilepsy diagnosis. However, significant challenges arise when a healthy PET control group is unattainable. Yaakub \emph{et al.} have previously introduced a Pix2PixGAN-based method for MRI to PET translation. This method used paired MRI and FDG PET scans from healthy individuals for training, and produced pseudo normal FDG PET images from patient MRIs that are subsequently used for lesion detection. However, this approach requires a large amount of high-quality, paired MRI and PET images from healthy control subjects, which may not always be available. In this study, we investigated unsupervised learning methods for unpaired MRI to PET translation for generating pseudo normal FDG PET for epileptic focus localization. Two deep learning methods, CycleGAN and SynDiff, were employed, and we found that diffusion-based method achieved improved performance in accurately localizing the epileptic focus.
\end{abstract}

\keywords{Medical Imaging Synthesis, Generative Adversarial Network, Diffusion Models}

\section{INTRODUCTION}
\label{sec:intro}  
Accurate localization of epileptic foci is critical for those who are drug-resistant and had to resort to surgical resection for seizure control. While routine MRI images are ubiquitously used for epilepsy diagnosis, a significant portion of them do not exhibit identifiable lesion (so-called MR negative cases) \cite{muhlhofer2017mri}. FDG PET is considered an important tool in lesion detection for focal epilepsy \cite{niu2021performance}. Conventionally, a healthy control group that consists of normal PET images is needed to improve diagnostic performance \cite{mayoral2016seizure}. Such control group may be used as visual check point, or as reference in statistical calculations (e.g. Z-statistics) \cite{zhang2022combined}. While it has been shown that closely matched healthy group yields better lesion detection \cite{de2018age}, acquisition of such control group remains challenging for many sites.  

Image-to-image translation is a common task in computer vision and has been recently performed for a wide range of applications in medical imaging. Deep learning-based methods that employ generative adversarial networks (GANs) have been actively investigated for performing cross modality medical image-to-image translation tasks. Li \emph{et al.} \cite{li2014deep} proposed a deep learning based convolutional neural network framework for estimating multi-modality imaging data in Alzheimer's Disease (AD) diagnosis. Pan \emph{et al.} \cite{pan2018synthesizing} used 3D Cycle-consistent Generative Adversarial Networks (3D-cGAN) to capture the underlying relationship between different modalities and complete missing PET, then developed a deep multi-instance neural network for AD diagnosis and prediction. Hu \emph{et al.} \cite{hu2021bidirectional} proposed a 3D end-to-end synthesis network BMGAN for brain MR-to-PET synthesis. Sikka \emph{et al.} \cite{sikka2018mri} estimated FDG PET scans from the given MR scans using a 3D U-Net architecture for improving classification accuracy. Recently, Yaakub \emph{et al.} \cite{yaakub2019pseudo} proposed the use of conditional GAN (cGAN) \cite{isola2017image} for synthesizing pseudo normal PET images from T1w MRI scans. The syntehsized pseudo normal PET images were subsequently employed as healthy control subjects and compared with the real PET scans to detect epilepsy. However, the cGAN used in that work needs to be trained on paired image data. For situations where paired image data are unavailable, the method proposed by Yaakub \emph{et al.} cannot be deployed.

To address the need for performing image-to-image translation tasks using only unpaired image data, Zhu \emph{et al.} proposed a Cycle-Consistent Adversarial Network (CycleGAN) method \cite{zhu2017unpaired}. Recently, denoising diffusion-based generative models \cite{ho2020denoising} have been proposed and achieved many successes in generating high-fidelity images that even possessed better image quality than GAN-generated images. More recently, an adversarial diffusion modeling method, SynDiff, has been proposed by {\"O}zbey \emph{et al.} to improve the performance of unsupervised medical image translation tasks \cite{ozbey2023unsupervised}.

In this work, we investigated the feasibility of SynDiff \cite{ozbey2023unsupervised} and CycleGAN \cite{zhu2017unpaired} to generate pseudo normal PET images given unpaired training image data for providing personalized control that could assist lesion localization for epilepsy patients. The image quality of the pseudo normal PET generated by the SynDiff and CycleGAN was objectively assessed. The generated pseudo normal PET images were subsequently used with the clinical PET scans (referred to hereafter as ``real PET'') to localize epilepic foci. The localization performance was assessed by comparing with clinical diagnosis results.

\section{Methods}
The proposed epileptic focus localization workflow consists of an image synthesis stage and a diagnosis stage. The image synthesis stage of the SynDiff \cite{ozbey2023unsupervised} is illustrated in Fig. \ref{workflow} and the training process of the SynDiff is illustrated in Fig. \ref{workflow_train}. Moreover, the diagnosis stage for localizing epileptic focus is illustrated in Fig. \ref{diagnosis}. More details are described below, and the details about the CycleGAN can be found in the publication \cite{zhu2017unpaired}.

\subsection{Adversarial Diffusion Models for Unpaired MR-to-PET Translation}

\begin{figure}[!ht]
    \centering
    \includegraphics[width=0.7\textwidth]{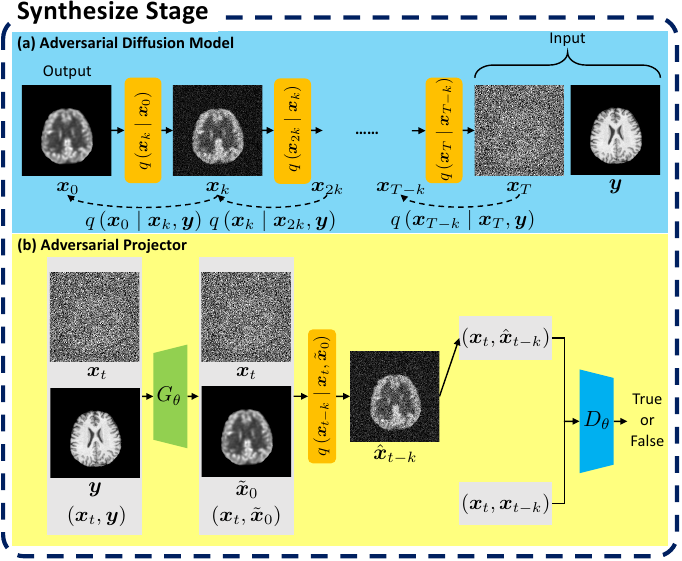}
    \caption{The proposed workflow for synthesizing pseudo normal PET images. It consists of two components: (a) adversarial diffusion model and (b) adversarial projector.}
    \label{workflow}
\end{figure}

SynDiff is a diffusion-based unsupervised method for medical image-to-image translation between unpaired source-target modalities. To synthesize pseudo normal PET, SynDiff performs fast diffusion forward steps. As shown in Fig. \ref{workflow} (a), it adds noise between $\boldsymbol{x}_0$ and $\boldsymbol{x}_T$ with a large step size, and employs the source MRI image $\boldsymbol{y}$ as conditional diffusion guidance for the reverse process. During the reverse process, SynDiff employs an adversarial projector, shown in Fig. \ref{workflow} (b), that allows for large diffusion steps for fast image sampling. This projector consists of a generator $G_\theta$ and a discriminator $D_\theta$. The generator $G_\theta$ first generates an estimate $\boldsymbol{\tilde{x}}_0$ of the denoised target PET image $\boldsymbol{x}_0$ based on the noisy target image $\boldsymbol{x}_t$ at time $t$ and the guidance MRI image $\boldsymbol{y}$. Subsequently, we can acquire denoised image samples $\hat{\boldsymbol{x}}_{t-k}$ synthesized from the denoising distribution $q\left(\boldsymbol{x}_{t-k} \mid \boldsymbol{x}_{t},\tilde{\boldsymbol{x}}_{0}\right)$. Simultaneously, the discriminator $D_\theta$ in the adversarial projector distinguishes between the tuples $\{\boldsymbol{x}_{t},\boldsymbol{x}_{t-k}\}$ and $\{\boldsymbol{x}_{t},\hat{\boldsymbol{x}}_{t-k}\}$. 

\begin{figure}[!ht]
    \centering
    \includegraphics[width=\textwidth]{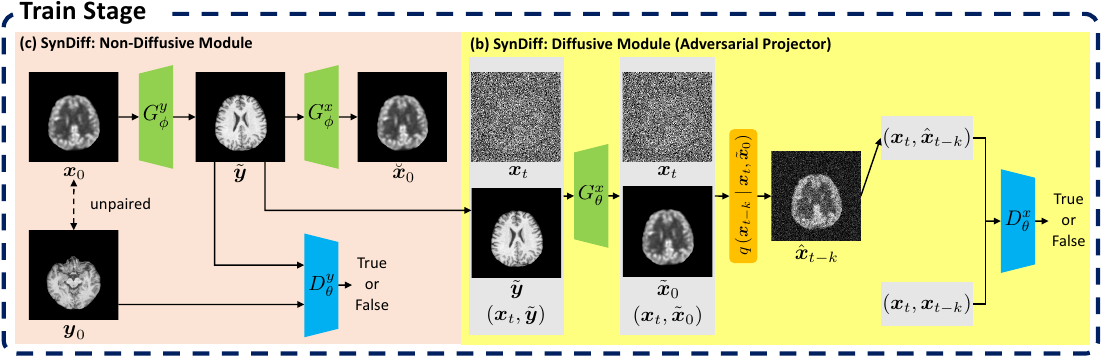}
    \caption{Training process of SynDiff. During training, each modality consists of (b) diffusive and (c) non-diffusive modules. Here only shows the diffusive and non-diffusive modules corresponding to the target modality $\boldsymbol{x}$ (PET), the process corresponding to the source modality $\boldsymbol{y}$ (MRI) are analogous.}
    \label{workflow_train}
\end{figure}

To ensure unsupervised learning in which paired images are unavailable, SynDiff adopts a non-diffusive model in the training process to generate estimated paired images for training the diffusive module, as shown in Fig. \ref{workflow_train}. Similar to CycleGAN, for each modality, the non-diffusive module has two generators $G_{\phi}^x$, $G_{\phi}^y$ and one discriminator $D_{\theta}^y$, aiming to estimate a source MRI image $\tilde{\boldsymbol{y}}$ given the corresponding target PET image $\boldsymbol{x}_{0}$, and the discriminator is responsible to distinguish real and generated images. Subsequently, $\tilde{\boldsymbol{y}}$ is used with $\boldsymbol{x}_{t}$, a noisy image of $\boldsymbol{x}_0$, in the training of the diffusive module. 
After training, real MRI image $\boldsymbol{y}$ is used for providing the guidance in the diffusive module (i.e., adversarial projector), as shown in Fig. \ref{workflow}. More specific details about the SynDiff model can be found in the publication \cite{ozbey2023unsupervised}.

\subsection{Synthesized Pseudo Normal PET Scans for Epileptic Focus Localization}

\begin{figure}[!ht]
    \centering
    \includegraphics[width=\textwidth]{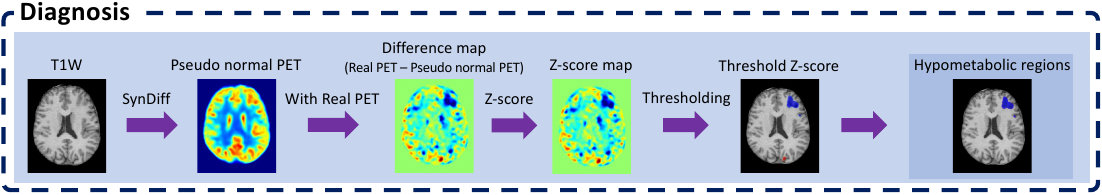}
    \caption{Illustration of pseudo normal PET assisted epileptogenic focus localization diagnosis workflow.}
    \label{diagnosis}
\end{figure}

To extract hypometabolic regions in epilepsy patients, we employed Z-score analysis that quantifies metabolic deviations from the pseudo normal PET. Negative Z-scores are interpreted as an indication of hypometabolism. As shown in Fig. \ref{diagnosis}, the difference map was generated by the real PET minus the pseudo normal PET generated by SynDiff or CycleGAN. It was subsequently converted to Z-score map as $\text{Z-score}=\frac{X-\mu}{\sigma}$, where $X$ was the pixel value in difference map, $\mu$ was the mean and $\sigma$ was the standard deviation across the whole difference map. To establish a threshold for detecting statistically significant hypometabolism, we opted for a one-sided hypothesis test, and set the significance level at $P=0.05$. By applying the norminv function within MATLAB 2022b (\href{https://www.mathworks.com/}{https://www.mathworks.com/}), we calculated a Z-score threshold: $Z<-1.65$ corresponded to our predetermined P-value. Meanwhile, we chose the cluster size threshold: $K>1500$ for detecting the lesion. All the calculations were limited in gray matter regions, which contained the neuronal cell bodies of the brain, serving as the principal source of cerebral electrical activity.

\section{Numerical Studies}
\subsection{Dataset and Implementation}
\subsubsection{Participants} This study has been approved by the Internal Review Board of Ruijin Hospital. Eighty patients diagnosed with refractory focal epilepsy between May 2018 and July 2022 were restrospectively included, and de-identified before data analysis. The diagnosis of these patients with epilepsy was obtained from medical records. All the patients had well documented lesion location based on video-EEG, routine MRI, FDG PET, and SEEG recordings. Additionally, 30 of them also had pre- and post-surgical MRI. Further, 36 healthy control subjects without any history of neurological or psychiatric disease were recruited, who underwent the same neuroimaging protocols of MRI and PET.

\subsubsection{Data Acquisition} PET and MRI scans were performed with an integrated 3.0-T hybrid PET/MR scanner (Biograph mMR; Siemens Healthcare, Erlangen, Germany). MRI sequences include 3D T1-weighted anatomical images using MPRAGE (resolution $0.5 \times 0.5 \times 1.0$ mm$^{3}$, TR/TE/TI 1,900/2.44/900 ms, FOV $250 \times 250$ mm$^{2}$, 192 slices), and T2-weighted FLAIR (resolution $0.4 \times 0.4 \times 3.0$ mm$^{3}$, TR/TE/TI 8,460/92/2433 ms, FOV $220 \times 220$ mm$^{2}$, 45 slices). All patients and controls were administered [$^{18}$F]FDG intravenously using a mean dose of 184.8 $\pm$ 29.0 MBq (range 133.2–247.9 MBq), and scanned 30-50 min after the injection. Static PET data were acquired in a sinogram mode for 15 min, having the size of $344 \times 344$, and post-filtered with an isotropic full-width half-maximum (FWHM) Gaussian kernel of 2 mm. Attenuation correction was performed using advanced PET attenuation correction with a unique 5-compartment model including bones \cite{koesters2016dixon}.

\subsubsection{Image Data Preprocessing} FreeSurfer v7.0 package (\href{https://surfer.nmr.mgh.harvard.edu}{https://surfer.nmr.mgh.harvard.edu}) was used to obtain processed T1-MPRAGE, which underwent motion correction and non-uniform intensity normalization. It was then processed by skull-stripping using SynthStrip tool \cite{hoopes2022synthstrip}. Each MRI was segmented into gray matter, white matter and cerebrospinal fluid using SPM12 software (\href{https://fil.ion.ucl.ac.uk/spm/}{https://fil.ion.ucl.ac.uk/spm/}) \cite{penny2011statistical}. [$^{18}$F]FDG PET were rigidly aligned to its corresponding MR images, and all ROI extraction of FDG PET were based on MRI segmentations. 

\subsubsection{Training and Generation of Pseudo Normal PET}
For model training, a total of 3,000 2D slice of unpaired MRI and PET images from 28 healthy subjects were used in this work, consisting of 1,500 MRI and 1,500 PET data. And we use 1,019 healthy paired MRI and PET from 8 healthy subjects for testing the MRI-to-PET image translation performance. Also, MRI from clinical scans of all the 80 patients were used to generate pseudo normal PET, which was subsequently used with paired real patient PET for the localization of epileptic focus task and the evaluation of localization performance. The data had the size of $256 \times 256$. SynDiff and CycleGAN were both trained on one NVIDIA GeForce RTX 3090 Ti GPU by use of PyTorch and Adam optimizer.

\subsection{Evaluation of Epileptic Focus Localization Performance}
Gold standard of epileptic focus localization were defined according to clinical reports. For all patients, epileptic focus had assignment to 8 major brain regions (4 lobes in each hemisphere: frontal, temporal, parietal, occipital) \cite{zhang2021high}.

We evaluated the performance of epileptic foci localization in the considered 80 patients. Detection rate was defined as the proportion of patients whose focus was identified. Localization accuracy was the proportion of identified patients whose estimated lesion location was in the brain region found in the clinical report.

\section{Results}
\begin{figure}[!ht]
    \centering
    \includegraphics[width=0.9\linewidth]{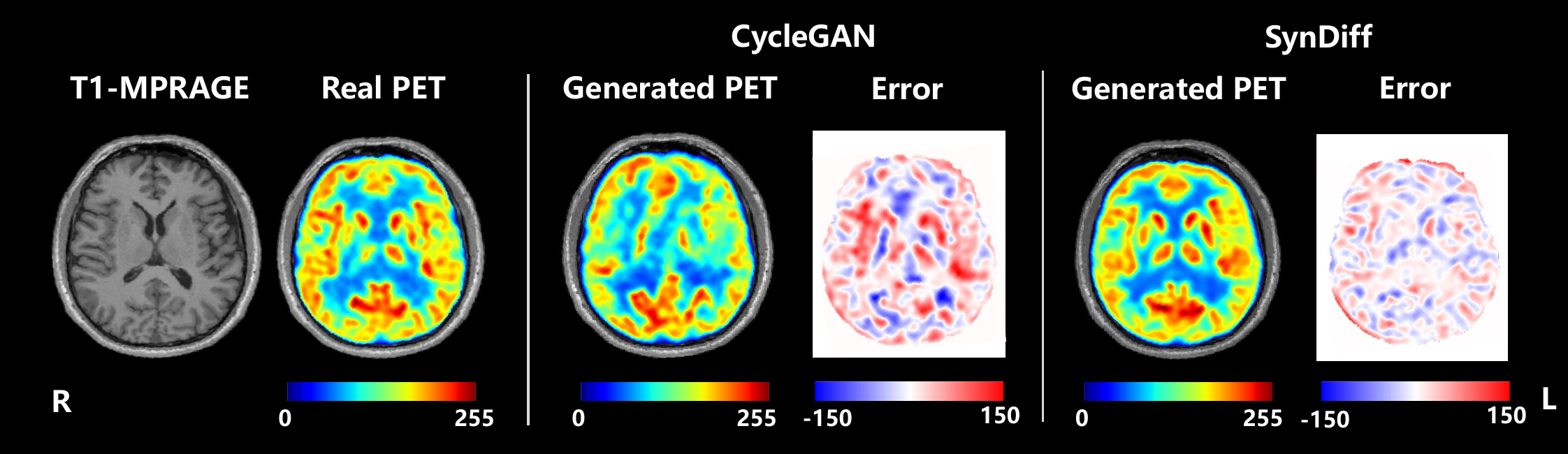}
    \caption{Results of MRI-to-PET image translation. Anatomical MR images and its corresponding real FDG PET image are shown on the left. Generated PET images by CycleGAN and SynDiff, as well as their corresponding error maps are displayed side-by-side.} 
    \label{fig:errormap}
\end{figure}

The SynDiff- and CycleGAN-generated PET images both look similar to real PET image (Figure \ref{fig:errormap}). However, as shown in Table. \ref{results}, the SynDiff model out-performed CycleGAN in terms of structural similarity (SSIM), Frechet inception distance (FID), peak signal-to-noise ratio (PSNR), and root mean squared error (RMSE).

\vspace{-4mm}
\begin{table}[!ht]
\centering
\begin{tabular}{ccccc}
\toprule
Method  & SSIM ($\uparrow$) & FID ($\downarrow$) & PSNR ($\uparrow$) & RMSE ($\downarrow$) \\ \midrule
CycleGAN (1,800 epoch) & 0.8100  & 70.8870 & 21.4471 & 0.0859 \\ 
SynDiff (70 epoch) & 0.9028 & 44.2106 & 24.3415 & 0.0616 \\ \bottomrule
\end{tabular}
\caption{Performance evaluation on test set.}
\label{results}
\end{table}
\vspace{-1mm}

\begin{figure}[!ht]
    \centering
    \includegraphics[width=0.6\linewidth]{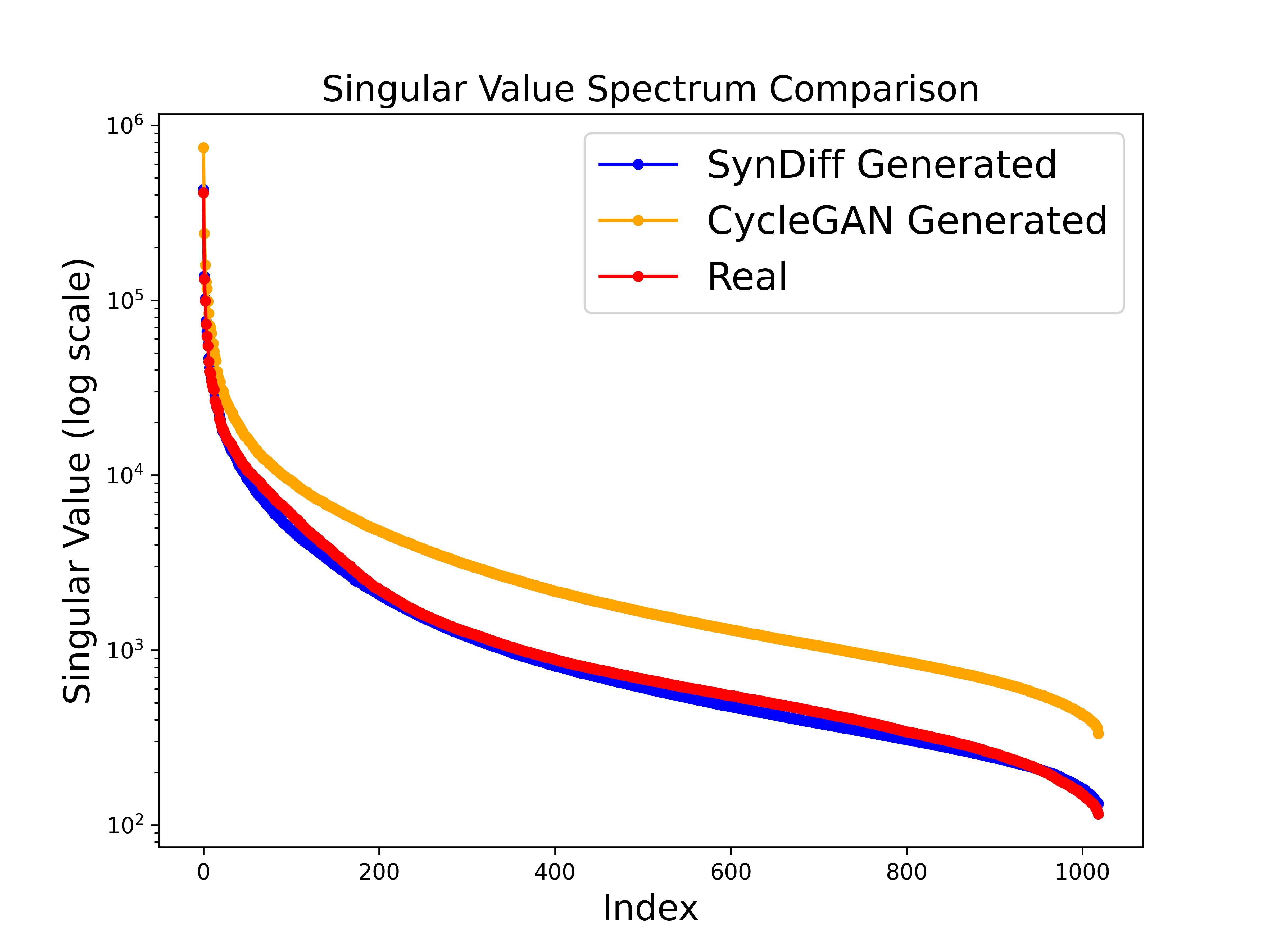}
    \caption{Singular value spectrum comparison between SynDiff- and CycleGAN-generated PET images and real PET images.}
    \label{svs}
\end{figure}

We also computed the singular value spectra for images produced by SynDiff and CycleGAN, and compare them to the ground truth in Fig. \ref{svs}. It is shown that the curve of SynDiff demonstrates a significantly greater overlap with the ground truth curve than the curve of CycleGAN.

\begin{figure}[!ht]
    \centering
    \includegraphics[width=0.9\linewidth]{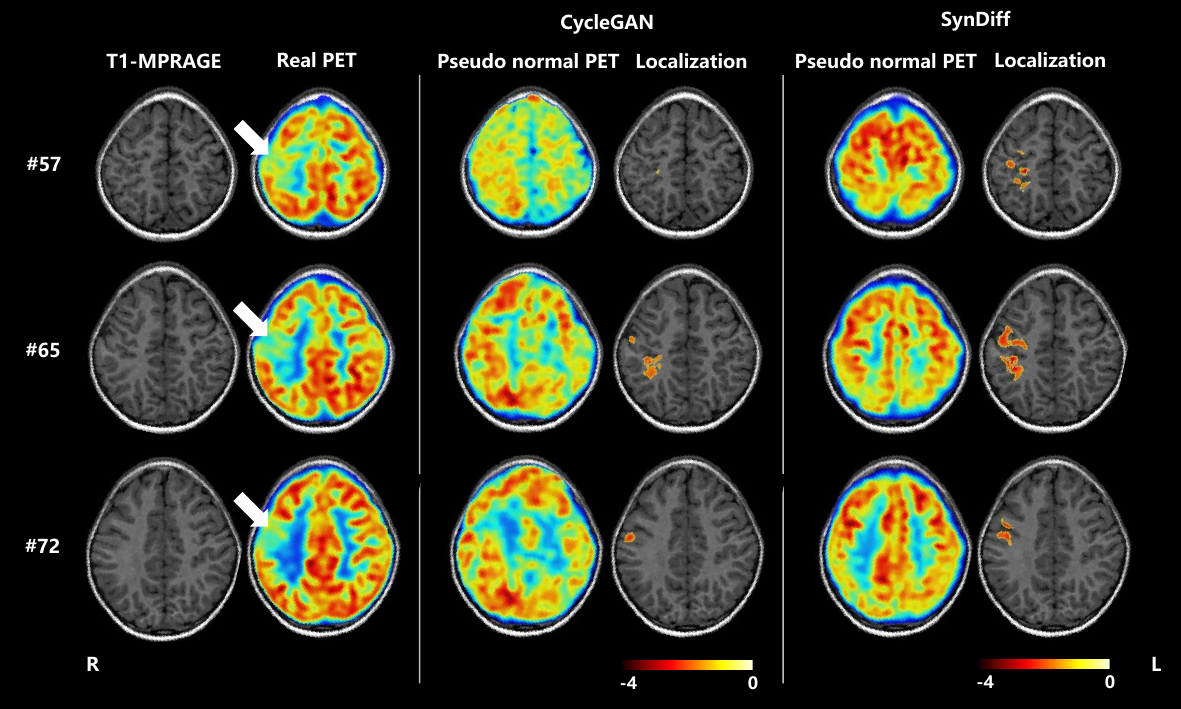}
    \caption{Representative case of epileptic patients. This figure showed three slices of case E028. The left column shows anatomical MRI and corresponding FDG PET images . The middle column and the right column shows pseudo normal PET generated by CycleGAN and SynDiff, as well as the epileptogenic focus localization overlayed on the anatomical MRI.}
    \label{fig:representativecase}
\end{figure}

Representative case of real PET from a patient and its corresponding pseudo normal PET generated by CycleGAN and SynDiff are shown in Fig. \ref{fig:representativecase}. Image structure is better preserved in pseudo normal PET images generated by SynDiff compared to those generated by CycleGAN.  

\begin{figure}[!ht]
    \centering
    \includegraphics[width=0.6\linewidth]{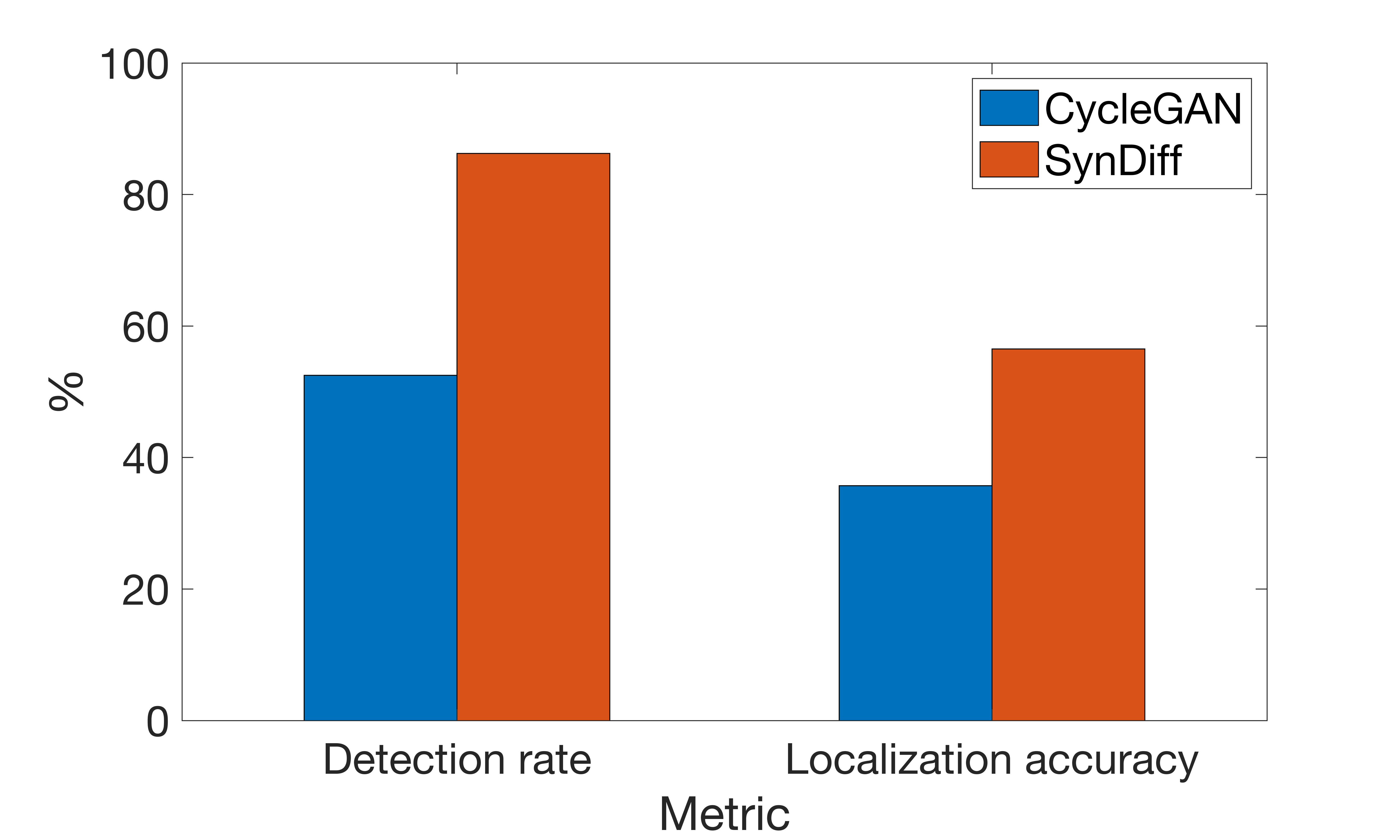}
    \caption{Comparison of the performance of pseudo normal PET generated by CycleGAN and SynDiff model for the localization of epileptic focus at patient level.}
    \label{bar}
\end{figure}

Fig. \ref{bar} summarized the performance of pseudo normal PET generated by CycleGAN and SynDiff model for the localization of epileptic focus. Pseudo normal PET images generated by SynDiff provide higher detection rate and localization accuracy than those generated by CycleGAN. 

\begin{figure}[!ht]
    \centering
    \includegraphics[width=0.9\linewidth]{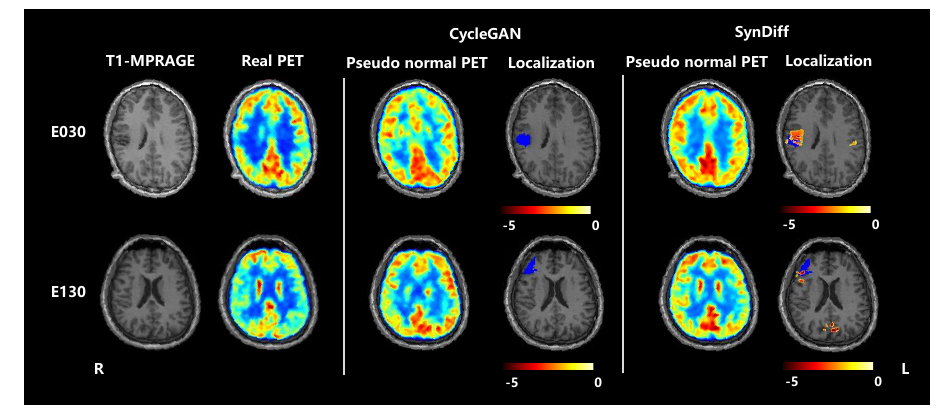}
    \caption{Illustration of pseudo normal PET assisted epileptogenic focus localization in challenging cases. The left column shows anatomical MRI and the corresponding FDG PET images of two cases. The middle column and the right column show pseudo normal PET generated by CycleGAN and SynDiff, as well as the epileptogenic focus localization overlayed on the anatomical MRI. Blue region represents surgical mask, red and yellow regions represent predicted epileptogenic focus.}
    \label{fig:challengingcases}
\end{figure}

Some patients represent challenging cases, where clinical reading of their PET does not provide conclusive diagnosis or lead to diagnostic errors. For such case, it is observed that SynDiff-produced pseudo normal PET can be useful for assisting the localization of suspicious regions, as shown in Fig. \ref{fig:challengingcases}.

\section{Conclusions}
In this study, we investigated the use of adversarial diffusion model SynDiff and CycleGAN to perform the unpaired MR-to-PET image translation task for localizing epilepic focus. Pseudo normal FDG PET images generated from patients' MRI scans were used as personalized control images for epileptic focus localization. It was demonstrated that the adversarial diffusion model outperformed CycleGAN to produce high-fidelity PET images used in healthy controls. Our findings confirmed that the proposed method can significantly bolster the efficacy of lesion localization in FDG PET. Together, these innovations pave the way for more accurate epileptic focus localization using PET imaging in the absence of high quality control data. 

\bibliography{report} 
\bibliographystyle{spiebib} 

\end{document}